\documentclass{article}
\usepackage{spconf,amsmath,graphicx,hyperref}
\usepackage[all]{hypcap}

\title{DEMO-POSE: DEPTH-MONOCULAR MODALITY FUSION FOR OBJECT POSE ESTIMATION}
\name{Rachit Agarwal$^{\star}$$^{\ddagger}$ \qquad Abhishek Joshi$^{\star}$$^{\ddagger}$ \qquad Sathish Chalasani$^{\star}$\thanks{$^{\ddagger}$Authors with equal contributions, Samsung R\&D Institute Bangalore} \qquad Woo Jin Kim$^{\dagger}$
}
\address{$^{\star}$Samsung R\&D Institute, Bangalore \\
$^{\dagger}$Samsung Electronics Suwon, Republic of Korea}
\begin{document}
\ninept
\maketitle
\begin{abstract}
Object pose estimation is a fundamental task in 3D vision with applications in robotics, AR/VR, and scene understanding. We address the challenge of category-level 9-DoF pose estimation (6D pose + 3D size) from RGB-D input, without relying on CAD models during inference. Existing depth-only methods achieve strong results but ignore semantic cues from RGB, while many RGB-D fusion models underperform due to suboptimal cross-modal fusion that fails to align semantic RGB cues with 3D geometric representations. We propose DeMo-Pose, a hybrid architecture that fuses monocular semantic features with depth-based graph convolutional representations via a novel multi-modal fusion strategy. To further improve geometric reasoning, we introduce a novel Mesh-Point Loss (MPL) that leverages mesh structure during training without adding inference overhead. Our approach achieves real-time inference and significantly improves over state-of-the-art methods across object categories, outperforming the strong GPV-Pose baseline by 3.2\% on 3D IoU and 11.1\% on pose accuracy on the REAL275 benchmark. The results highlight the effectiveness of depth–RGB fusion and geometry-aware learning, enabling robust category-level 3D pose estimation for real-world applications.
\end{abstract}

\section{Introduction}
\label{sec:intro}

\begin{figure}[t]
    \vspace{0.2cm}
    \centering
    \includegraphics[width=1.0\linewidth]{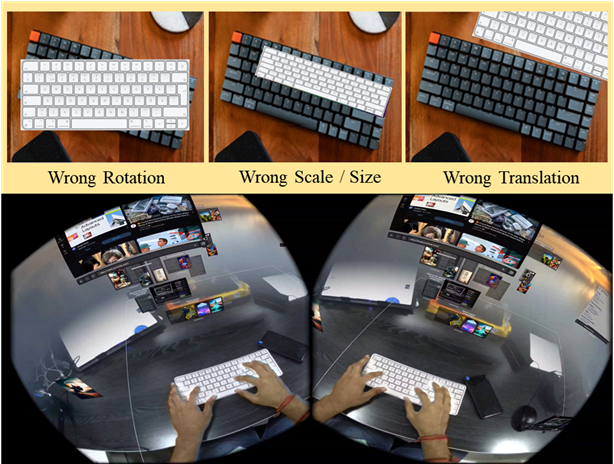}\label{fig:keyb_wrongR}\hfill  
    \vspace{-0.2cm}
    \caption{Impact of inaccurate rotation, scale, and translation when overlaying a virtual keyboard template on a physical keyboard in MR. The bottom image shows correct alignment using pose estimation for improved user experience, as seen through the HMD~\cite{Meta_Quest}
    }
    \label{fig:usecase}
    \vspace{-0.2cm}
\end{figure}

Accurate estimation of the \textbf{9-DoF object pose}---comprising 3D position, orientation, and absolute size---is a fundamental problem in computer vision. Robust pose prediction enables critical applications in \textit{robotic manipulation, autonomous navigation, and AR/VR scene understanding}, where reliable 3D reasoning is essential. 3D pose estimation is widely used in AR/VR with HMDs, smartphones, and gaming. However, inaccurate 9-DoF pose (rotation, translation, size) degrades AR user experience and poses risks in critical tasks like autonomous navigation, highlighting the need for robust pose estimation, as shown in Fig.~\ref{fig:usecase}. Pose estimation remains an active research domain in the computer vision community, with recent methods that can reliably estimate the pose even under severe occlusion~\cite{Di_2021_ICCV}.

While significant progress has been made in \textbf{6D pose estimation}, most existing approaches remain limited to \textit{instance-level settings}, often requiring precise CAD models at inference. Such methods fail to generalize to \textit{category-level pose estimation}, where unseen objects from known categories must be localized and scaled. Recent works have explored this problem, but \textbf{depth-only approaches typically outperform RGB-D methods}, indicating that current RGB--Depth fusion strategies are suboptimal.

\textbf{Instance-level pose estimation:}  
Significant progress has been made in \textbf{6D instance-level pose estimation}, where methods predict the pose of known objects with CAD models available. Existing approaches include direct regression~\cite{kehl2017ssd, hu2020single}, learning latent embedding for pose retrieval~\cite{tian2020shape}, correspondence-based methods with Perspective-n-Point (PnP) solvers~\cite{park2019pix2pose}, and fusion-based techniques~\cite{he2021ffb6d}. Despite strong performance, these methods are limited in practice: they require precise CAD models and typically handle only a small set of object instances.

\textbf{Category-level pose estimation:}  
Category-level methods aim to generalize pose estimation to unseen objects from known categories, without relying on CAD models. Early works introduced canonical spaces such as NOCS~\cite{wang2019normalized} and its extensions~\cite{chen2020cass, lin2021donet}, while later approaches incorporated geometric priors or dual networks~\cite{chen2021sgpa, lin2021dualposenet}. Depth-only methods have shown superior accuracy compared to RGB-D fusion~\cite{di2022gpv}, highlighting a gap in leveraging semantic cues from RGB images. Recent self-supervised strategies~\cite{peng2022self, lunayach2024fsd} reduce annotation cost but still lag behind supervised baselines in accuracy and efficiency.  

In summary, prior work either sacrifices generalization (instance-level) or fails to exploit RGB--Depth complementarity effectively (category-level). Our work addresses this gap by \textbf{fusing monocular RGB features with depth-based representations}, coupled with a novel geometry-aware loss, to achieve robust and real-time category-level 9-DoF pose estimation.

\begin{figure*}[th!]
\begin{center}
% \vspace{-2mm}
\includegraphics[width=\linewidth]{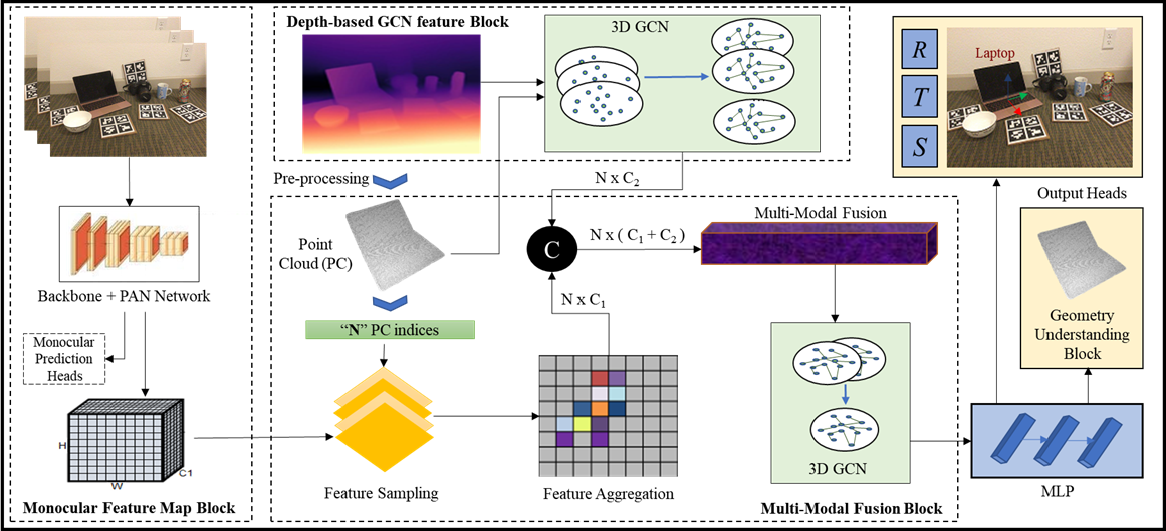}

\end{center}
\vspace{-0.7cm}
\caption{Fusion module architecture to leverage RGB features obtained from Monocular detection model and fuse with Depth-based GCN features. In inference, we achieve real-time performance, suitable for device deployment. We provide more details  in proposed method sec.\ref{sec:proposed_method}}
\label{fig:fusion}
\vspace{-0.2cm}
\end{figure*}

In this work, we introduce \textbf{DeMo-Pose}, a hybrid framework for category-level 9-DoF object pose estimation that overcomes these limitations. Our model fuses \textbf{monocular RGB features} with \textbf{depth-based graph convolutional representations} through a novel multi-modal fusion scheme. Furthermore, we propose a \textbf{Mesh-Point Loss (MPL)} that leverages 3D mesh structure during training to improve geometric reasoning without adding inference overhead.

\textbf{Our key contributions are as follows.}
\begin{itemize}
    \item A \textbf{hybrid fusion architecture} that integrates monocular semantic cues with depth features for pose estimation.
    \item A \textbf{geometry-aware training objective (Mesh Point Loss, MPL)} that enhances category-level generalization.
    \item A \textbf{real-time system (frames per second, FPS $\approx$18)}, achieving significant improvements over state-of-the-art methods, with \textbf{3.2\% gain in 3D-IoU} and \textbf{11.1\% gain in pose accuracy} on the REAL275 benchmark.
\end{itemize}

These contributions highlight the potential of RGB--Depth fusion and geometry-aware learning for advancing category-level 3D pose estimation in practical, real-world environments.

\begin{figure*}[th!]
\begin{center}
% \vspace{-2mm}
\includegraphics[width=\linewidth]{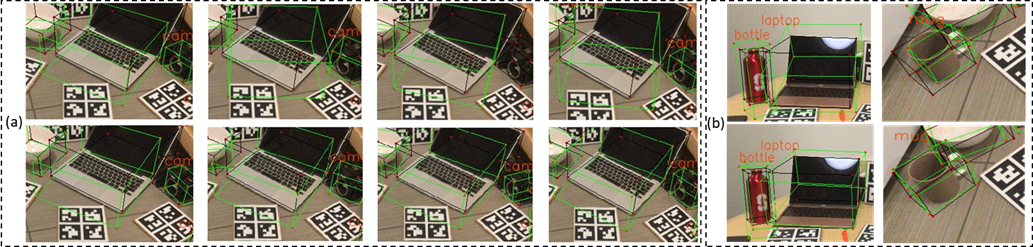}

\end{center}
\vspace{-0.7cm}
\caption{\textbf{(a) Comparison of predictions across video frames}: GPV-Pose exhibits temporal instability for the laptop category (top row), while our DeMo-Pose fusion yields stable predictions (bottom row). \textbf{(b) We represent predictions with green colored boxes and ground truth with black boxes}, our method (bottom row) produces tighter and more accurate boxes for laptop and mug compared to GPV-Pose (top row).}
\label{fig:subjective_results}
\vspace{-0.2cm}
\end{figure*}

\section{Proposed Method}
\label{sec:proposed_method}

We propose \textbf{DeMo-Pose},  a novel hybrid architecture that fuses semantic information derived from RGB input with features obtained from depth-based pose estimation model. The overall pipeline is shown in Fig.~\ref{fig:fusion}. Given an RGB image as input, we propose a single-stage detector architecture that can predict projected 3D-keypoints and relative size of objects. The details are provided in Section~\ref{ssec:mono}. The network learns pose-rich semantic and object-related cues through training in an end-to-end manner. We leverage these important spatial features as described in Section~\ref{ssec:depth} to accurately predict and improve the depth-based pose estimation model. To further improve geometry reasoning, we introduce in Section~\ref{ssec:mpl}, a \textbf{Mesh-Point Loss (MPL)} that leverages mesh structure during training without increasing inference cost.  

\subsection{Monocular Detection}
\label{ssec:mono}
Given an RGB input, our monocular method targets category-level 3D object pose estimation. Unlike approaches such as CenterPose~\cite{lin2022single}, which employ separate networks per category, we adopt a generic single-stage detector scalable across object classes. Following prior works~\cite{kehl2017ssd, chen2021sgpa}, we predict 2D projections of 3D cuboid corners and apply a Perspective-n-Point (PnP) algorithm to recover object pose. We leverage FCOS~\cite{tian2019fcos} backbone and pyramid features to predict 2D keypoints, class labels, and relative object size. Training explicitly exploits RGB semantics to learn rich pose cues by predicting relative object dimensions instead of absolute depth, thereby addressing the ill-posed nature of monocular depth estimation~\cite{lin2022single}. To avoid this, we estimate the relative dimensions that serve as robust auxiliary cues for downstream fusion.

\textbf{Monocular Architecture:} As shown in Fig.~\ref{fig:fusion}, the monocular network takes RGB images as input. It consists of a Backbone, Path Aggregation Network (PAN), and three Monocular Prediction Heads. We use GhostNet~\cite{Han2019GhostNetMF} as the backbone, where the Ghost module efficiently generates feature maps through inexpensive linear transformations, making it a strong choice for embedded devices with limited resources due to its balance of accuracy and efficiency. The multi-level backbone features are passed through a PAN, which enhances information flow via bottom-up augmentation, reducing the path between low- and high-level features.

Following PAN, three prediction heads jointly perform: i) regression of 8 projected 2D keypoints of a cuboid enclosing the object, ii) classification of the object, and iii) regression of its relative dimensions. Inspired by FCOS~\cite{tian2019fcos} , these heads share stages for efficient learning. The network is trained end-to-end with GIoU loss, quality focal loss, and distributional focal loss~\cite{li2020generalized_QFL_DFL}, commonly used in 2D detection and adapted here for monocular 3D detection, where estimated cuboid keypoints also act as projected 2D points.

Once the monocular model is trained, we freeze it and use the semantically rich PAN features as the Monocular Feature Map, for fusion to build the hybrid model. The second training phase of the hybrid architecture is detailed in the next section.

\subsection{Depth-based Backbone and Fusion}
\label{ssec:depth}
\vspace{-0.5em}
Depth information is essential for 9-DOF pose estimation. Recently, 3D graph convolution (3DGC)~\cite{lin2020convolution} has gained popularity due to its robustness to point cloud shift and scale. GPV-Pose~\cite{di2022gpv} employs 3DGC as a backbone to extract global and local features, enabling confidence-driven closed-form 3D rotation recovery and class-specific geometric characterization. It achieves $\approx$20 FPS on standard benchmarks, making it an effective depth-based baseline for our Depth–Monocular fusion approach.

Following~\cite{di2022gpv}, we preprocess depth maps with Mask-RCNN~\cite{he2017mask} to segment objects, back-project to 3D, and sample 1028 points as GPV-Pose input. The 3DGC extracts global and per-point features, which are fed to regression heads to predict pose \{r, t, s\} (Rotation, Translation, Size). However, depth-only features are sensitive to noise, occlusion, sampling, and segmentation. In contrast, analogous to scene-understanding tasks, RGB images provide contextual and semantic cues (e.g., occlusion and background information) useful for pose estimation. Hence, we fuse RGB and depth modalities to exploit their complementary strengths.

As shown in Fig.~\ref{fig:fusion}, monocular features (H×W×C$_1$) and depth-based features (N×C$_2$) lie in different spaces. We align them via a feature sampling module: using indices of N sampled point cloud points, their spatial locations are bilinearly interpolated on the monocular feature map to form an N×C$_1$ tensor. This makes features dimensionally compatible with depth-based features for fusion. While we adopt concatenation for simplicity, our approach generalizes to other fusion strategies such as addition, Hadamard multiplication, or MLP-based fusion.

\vspace{-0.7em}
\subsection{Geometry-Aware Mesh-Point Loss (MPL)}
\label{ssec:mpl}
\vspace{-0.5em}
To further regularize pose estimation, we introduce the \textbf{Mesh-Point Loss (MPL)}. During training, a subset of vertices is sampled from the ground-truth object mesh using Poisson disk sampling~\cite{yuksel2015poisson}. The network predicts corresponding mesh points, and the L2 distance between predicted and ground-truth vertices is minimized:
\begin{equation}
    \mathcal{L}_{MPL} = \frac{1}{V} \sum_{i=1}^{V} \| R \cdot M^{GT}_i - M^{pred}_i \| ^2 ,
\end{equation}
where $M^{GT}_i$ and $M^{pred}_i$ are ground-truth and predicted vertices, and $R$ is the ground-truth rotation.  

Our MPL is inspired from PoseLoss proposed in PoseCNN~\cite{xiang2017posecnn}. PoseLoss deals with rotation matrix computed from estimated quaternion, whereas, we directly regress the 3D mesh vertices. Unlike PoseLoss, MPL can handle object symmetries since it directly supervises 3D point distributions rather than rotation parameters. Thus, MPL overcomes the inherent issues in PoseLoss. Since mesh predictions are only required during training, MPL introduces no overhead at inference. The total loss combines $L_{base}$ with MPL:
\begin{equation}
    \mathcal{L}_{total} = \mathcal{L}_{base} +  \lambda_{MPL} *  \mathcal{L}_{MPL} 
\end{equation}
where $\lambda_{MPL}$ is a hyperparameter, and, $L_{base}$ denotes the standard GPV-Pose regression loss supervising rotation, translation and scale. 

The proposed architecture achieves three desirable properties:  
(i) effective fusion of semantic RGB and depth-based geometry,  
(ii) geometry-aware training via MPL, and  
(iii) efficient inference, as monocular features can also be used standalone in resource-constrained scenarios.

\section{Experiments and Results}
\label{sec:experiments}

\subsection{Implementation Details}
\label{ssec:implementation}
To ensure fair benchmarking, we train our method purely on real data, following~\cite{tian2020shape, chen2021sgpa, di2022gpv}, and generate instance masks using Mask-RCNN~\cite{he2017mask}. Adopting the GPV-Pose setup, 1028 points are uniformly sampled from the back-projected depth map as input to the Depth-based model. Multi-modal fused features are trained with existing losses using default hyperparameters. For Mesh-Point Loss (MPL), we introduce a scalar weight $\lambda_{\text{MPL}}$ to balance gradient magnitudes, empirically set to 2000. Training is conducted in PyTorch on the REAL275 dataset with a single model across all categories. Details of hyperparameter selection experiments will be provided in the appendix. For consistency, we adopt the same hardware settings as reported in GPV-Pose.
\vspace{-0.1cm}

\subsection{Dataset and Evaluation Metrics}
\label{ssec:dataset}
We evaluate DeMo-Pose on the widely used \textbf{REAL275} benchmark~\cite{wang2019normalized}, which consists of challenging 13 real-world scenes covering six object categories. i.e. \textit{bottle, bowl, camera, can, laptop, and mug}. Following standard protocol, 7 scenes ($\approx$4.3k images) are used for training and 6 scenes ($\approx$2.7k images) for testing. The proposed model leads to superior results when benchmarked against prior pose estimation models as shown in Table \ref{table:sota_result} and present subjective comparison in Fig.~\ref{fig:subjective_results}.

We follow prior works~\cite{chen2021sgpa, di2022gpv} and report:  
(i) \textbf{3D IoU} at 25\%, 50\% and 75\% thresholds (3D$_{25}$, 3D$_{50}$, 3D$_{75}$) to evaluate joint translation, rotation, and size accuracy, and  
(ii) \textbf{pose accuracy} under combined rotation and translation thresholds, including 5\textdegree2cm, 5\textdegree5cm, 10 \textdegree5cm, and 10\textdegree10cm, and (iii) \textbf{frame rate} (FPS) for each method. Our method achieves efficient inference, following the GPV-Pose implementation settings.
\vspace{-0.1cm}

\begin{table*}[h]
\centering
\begin{tabular}{l c | c | c c c | c c c c | c}
\hline
\textbf{Method} & & \textbf{Setting} & {\textbf{3D$_{25}$}}$\uparrow$ & {\textbf{3D$_{50}$}}$\uparrow$ & {\textbf{3D$_{75}$}}$\uparrow$ & {\textbf{5\textdegree2cm}}$\uparrow$ & {\textbf{5\textdegree5cm}}$\uparrow$ & {\textbf{10\textdegree5cm}}$\uparrow$ & {\textbf{10\textdegree10cm}}$\uparrow$ &
{\textbf{FPS}} \\ 
\hline \hline
NOCS~\cite{wang2019normalized} & CVPR'19  & RGB-D & \textbf{84.9} & 80.5 & 30.1 & 7.2 & 10.0 & 25.2 & 26.7 & 5 \\
CASS~\cite{chen2020cass} &  CVPR'20 & RGB-D & 84.2 & 77.7 & - & - & 23.5 & 58.0 & 58.3 & - \\
SPD~\cite{tian2020shape} & ECCV'20 & RGB-D & 83.4  & 77.3 & 53.2 & 19.3 & 21.4 & 54.1 & - & 4 \\
CR-Net~\cite{wang2021category} & IROS'21   & RGB-D & - & 79.3 & 55.9 & 27.8 & 34.3 & 60.8 & - & - \\
SGPA~\cite{chen2021sgpa} & ICCV'21  & RGB-D & - & 80.1 & 61.9 & 35.9 & 39.6 & 70.7 & - & - \\
DualPoseNet~\cite{lin2021dualposenet} & ICCV'21 & RGB-D & -  & 79.8 & 62.2 & 29.3 & 35.9 & 66.8 & - & 2 \\

DO-Net~\cite{lin2021donet} & Arxiv'21 & D & - & 80.4 & 63.7 & 24.1 & 34.8 & 67.4 & - & 10 \\
FS-Net~\cite{chen2021fs} & CVPR'21 & D & - & - & - & - & 28.2 & 60.8 & 64.6 &20  \\       
GPV-Pose~\cite{di2022gpv} & CVPR'22  & D & 84.2 & 83.0 & 64.4 & 32.0 & 42.9 & 73.3 & 74.6 & 20 \\

SSC-6D~\cite{peng2022self} & AAAI'22 & RGB-D & 83.2 & 73.0 & - & - & 19.6 & 54.5 & 64.4 & - \\

FSD~\cite{lunayach2024fsd} & ICRA'24 & RGB-D & 80.9 & 77.4 & - & - & 28.1 & 61.5 & 72.6 & - \\
DiffusionNOCS~\cite{ikeda2024diffusionnocs} & IROS'24 & RGB-D & - & - & - & - & 35.0 & 66.6 & - & - \\

ACR-Pose-PN2~\cite{fan2024acr} & ICMR'24 & RGB-D & - & 82.3 & 66.6 & \textbf{36.7} & 41.3 & 56.7 & 67.0 & - \\
GS-Pose~\cite{wang2024gs} & ECCV'24 & RGB-D & 82.1 & 63.2 & - & - & 28.8 & - & 60.6 & - \\
                    \hline
% Ours (w/o MPL) & & RGB-D & 80.5 & 82.8 & 64.1 & 33.6 & 46.3 & 78.8 & 80.1 & 17.86 \\                     
\textbf{DeMo-Pose (Ours)} & & RGB-D & 84.2 & \textbf{83.0} & \textbf{66.8} & 34.7 & \textbf{47.7} & \textbf{79.3} & \textbf{80.6} & 17.86 \\      
                    \hline
\end{tabular}
\vspace{-0.1cm}
\caption{Comparison with state-of-the-art methods on the REAL275 dataset. \textbf{Bold} indicates superior results. $\uparrow$ denotes higher is better for each metric. } % and the second best results are \underline{underlined}.}
\label{table:sota_result}
\end{table*}

\begin{table}[t]
\centering
\begin{tabular}{lcccc}
\hline
\hspace{-0.1cm}\textbf{Method} & {\textbf{3D$_{75}$}} & {\textbf{5\textdegree2cm}} & {\textbf{5\textdegree5cm}} & {\textbf{10\textdegree5cm}}\hspace{-0.1cm} \\
\hline
\hspace{-0.1cm}DeMo-Pose w/o MPL & 64.1 & 33.6 & 46.3 & 78.8 \\
\hspace{-0.1cm}DeMo-Pose + MPL   & \textbf{66.8} & \textbf{34.7} & \textbf{47.7} & \textbf{79.3} \\
\hline
\end{tabular}
\vspace{-0.1cm}
\caption{Ablation study on efficacy of our Mesh Point Loss (MPL) on REAL275 test set}
\label{table:ablationMPL}
\end{table}

\begin{table}[t]
\centering
\begin{tabular}{lccccc}
\hline
\hspace{-0.1cm}\textbf{Fusion Methods} & {\textbf{3D$_{50}$}} & {\textbf{3D$_{75}$}} & {\textbf{5\textdegree5cm}} & {\textbf{10\textdegree5cm}} & \hspace{-0.1cm}{\textbf{FPS}}\hspace{-0.1cm} \\
\hline
\hspace{-0.1cm}Concatenation & \textbf{83.0} & 66.8 & \textbf{47.7} & \textbf{79.3} & \hspace{-0.2cm}\textbf{17.86}\hspace{-0.1cm} \\
\hspace{-0.1cm}MLP + Skip   & 82.8 & 66.2 & 46.0 & 77.6 & \hspace{-0.2cm}17.41\hspace{-0.1cm} \\
\hspace{-0.1cm}Attention + Skip   & 82.9 & \textbf{66.9} & 46.8 & 77.9 & \hspace{-0.2cm}17.36\hspace{-0.1cm} \\
\hline
\end{tabular}
\vspace{-0.1cm}
\caption{Ablation study on RGB–D fusion strategies}
\vspace{-0.2cm}
\label{table:rgbd_fusion}
\end{table}

\subsection{Results}
\label{ssec:results}
\vspace{-0.1cm}
We compare our method with state-of-the-art models on REAL275 dataset. Table \ref{table:sota_result} compares DeMo-Pose with representative instance and category-level baselines. Our method achieves consistent improvements across most metrics (i.e., 5 out of 6 metrics). Notably, DeMo-Pose surpasses the strong depth-only baseline GPV-Pose by \textbf{3.2\% on 3D75 IoU} and \textbf{11.1\% on 5\textdegree5cm pose accuracy}. Furthermore, for the \textbf{10\textdegree5cm, 10\textdegree10cm metrics}, we surpass the prior art by a relative increase of \textbf{8.1\% and 7.4\%} , whilst running almost in real time, \textbf{FPS $\approx$18}. These results validate the benefit of fusing semantic RGB cues with depth-based geometric features. In the appendix, we will provide a detailed comparison of the per-category results of our method on REAL275. 

For better understanding, we provide qualitative comparisons (not shown for brevity), indicating that our fusion strategy produces more stable predictions, compared to GPV-Pose. In Fig.~\ref{fig:subjective_results}(a), for a given sequence of frames in a video, it is observed that the predictions flicker for the laptop category in GPV-Pose, while with our DeMo-Pose fusion approach, the predictions exhibit improved temporal consistency and stability. Furthermore, Fig.~\ref{fig:subjective_results}(b) shows our method produces tighter and more accurate boxes for laptop and mug. Overall, the results highlight that: (i) RGB cues provide complementary semantics missing in depth-only pipelines, (ii) MPL improves geometry awareness and robustness, and (iii) the hybrid model achieves real-time inference, making it suitable for deployment in AR/VR and robotics. Next, we provide ablation studies on novel Mesh-Point Loss and RGB-D fusion strategies, with objective comparison results --- strongly confirming the efficacy of our proposed approach.
\vspace{-0.1cm}

\subsection{Ablation on Mesh-Point Loss (MPL)}
\label{ssec:mpl_ablation}
\vspace{-0.1cm}
To understand the efficacy of the proposed Mesh-Point loss (MPL), we analyze the performance through an ablation study. In Table~\ref{table:ablationMPL}, we refer to our baseline model i.e. Depth-Monocular based fusion as DeMo-Pose.  Adding MPL consistently improves performance across all metrics, demonstrating that explicitly supervising geometry strengthens category-level pose prediction.

\vspace{-0.1cm}
\subsection{Ablation on RGB-D fusion strategies}
\label{ssec:fusion_ablation}
\vspace{-0.1cm}
To validate the contribution of the RGB--D fusion module, we conducted an ablation study on different fusion mechanisms. Specifically, we evaluate three strategies: (i) \textbf{Concatenation}, where RGB and depth features are directly concatenated; (ii) \textbf{MLP-based fusion with skip connection}, where features are projected into a common space and combined through a multilayer perceptron; and (iii) \textbf{Attention-based fusion with skip connection}, where cross-modal attention is used to adaptively weigh RGB and depth features. The results are summarized in Table~\ref{table:rgbd_fusion}. The results demonstrate that simple concatenation remains the most effective strategy for RGB-D integration in the proposed framework, without introducing any additional complexity.

\vspace{-0.1cm}
\section{Conclusion}
\label{ssec:conclusion}
\vspace{-0.1cm}
We presented \textbf{DeMo-Pose}, a hybrid framework for category-level 9-DoF object pose estimation that fuses semantic RGB features with depth-based geometric representations. By introducing a novel \textbf{Mesh-Point Loss (MPL)}, our method strengthens geometry awareness during training without adding inference overhead.  

Extensive experiments on the REAL275 benchmark demonstrate that DeMo-Pose achieves \textbf{state-of-the-art performance}, surpassing strong depth-only baselines by \textbf{3.2\% on 3D IoU} and \textbf{11.1\% on pose accuracy}, while maintaining efficient inference. These results highlight the effectiveness of multi-modal fusion and geometry-aware training for robust 3D vision.  

In future work, we plan to extend DeMo-Pose towards generative frameworks and large-scale datasets for holistic 3D scene understanding, enabling broader applications in AR/VR and robotics.


\bibliographystyle{IEEEbib}
\bibliography{strings,refs}

@online{Meta_Quest,
  author = {Meta Quest Pro},
  title = {https://www.meta.com/quest/quest-pro/},
  year = {2023},
  url = {https://www.meta.com/quest/quest-pro/},
  urldate = {}
}

@inproceedings{Di_2021_ICCV,
author = {Di, Yan and Manhardt, Fabian and Wang, Gu and Ji, Xiangyang and Navab, Nassir and Tombari, Federico},
title = {SO-Pose: Exploiting Self-Occlusion for Direct 6D Pose Estimation},
booktitle = {Proceedings of the IEEE/CVF International Conference on Computer Vision (ICCV)},
month = {October},
year = {2021},
pages = {12396-12405}
}

@inproceedings{lin2022single,
  title={Single-stage keypoint-based category-level object pose estimation from an RGB image},
  author={Lin et al., Yunzhi},
  booktitle={2022 International Conference on Robotics and Automation (ICRA)},
  pages={1547--1553},
  year={2022},
  organization={IEEE}
}

@article{lin2021donet,
  title={Donet: Learning category-level 6d object pose and size estimation from depth observation},
  author={Lin, Haitao and Liu, Zichang and Cheang, Chilam and Zhang, Lingwei and Fu, Yanwei and Xue, Xiangyang},
  journal={arXiv preprint arXiv:2106.14193},
  volume={2},
  number={4},
  year={2021}
}

@inproceedings{di2022gpv,
  title={Gpv-pose: Category-level object pose estimation via geometry-guided point-wise voting},
  author={Di et al., Yan},
  booktitle={Proceedings of the IEEE/CVF Conference on Computer Vision and Pattern Recognition},
  pages={6781--6791},
  year={2022}
}

@inproceedings{tian2020shape,
  title={Shape prior deformation for categorical 6d object pose and size estimation},
  author={Tian, Meng and Ang, Marcelo H and Lee, Gim Hee},
  booktitle={Computer Vision--ECCV 2020: 16th European Conference, Glasgow, UK, August 23--28, 2020, Proceedings, Part XXI 16},
  pages={530--546},
  year={2020},
  organization={Springer}
}

@inproceedings{kehl2017ssd,
  title={Ssd-6d: Making rgb-based 3d detection and 6d pose estimation great again},
  author={Kehl, Wadim and Manhardt, Fabian and Tombari, Federico and Ilic, Slobodan},
  booktitle={Proceedings of the IEEE international conference on computer vision},
  pages={1521--1529},
  year={2017}
}

@inproceedings{hu2020single,
  title={Single-stage 6d object pose estimation},
  author={Hu, Yinlin and Fua, Pascal and Wang, Wei and Salzmann, Mathieu},
  booktitle={Proceedings of the IEEE/CVF conference on CVPR},
  pages={2930--2939},
  year={2020}
}

@inproceedings{park2019pix2pose,
  title={Pix2pose: Pixel-wise coordinate regression of objects for 6d pose estimation},
  author={Park, Kiru and Patten, Timothy and Vincze, Markus},
  booktitle={Proceedings of the IEEE/CVF international conference on computer vision},
  pages={7668--7677},
  year={2019}
}

@inproceedings{he2021ffb6d,
  title={Ffb6d: A full flow bidirectional fusion network for 6d pose estimation},
  author={He, Yisheng and Huang, Haibin and Fan, Haoqiang and Chen, Qifeng and Sun, Jian},
  booktitle={Proceedings of the IEEE/CVF conference on CVPR},
  pages={3003--3013},
  year={2021}
}

@inproceedings{wang2019normalized,
  title={Normalized object coordinate space for category-level 6d object pose and size estimation},
  author={Wang, He and Sridhar, Srinath and Huang, Jingwei and Valentin, Julien and Song, Shuran and Guibas, Leonidas J},
  booktitle={Proceedings of the IEEE/CVF Conference on Computer Vision and Pattern Recognition},
  pages={2642--2651},
  year={2019}
}

@inproceedings{chen2020cass,
  title={Learning canonical shape space for category-level 6d object pose and size estimation},
  author={Chen, Dengsheng and Li, Jun and Wang, Zheng and Xu, Kai},
  booktitle={Proceedings of the IEEE/CVF conference on CVPR},
  pages={11973--11982},
  year={2020}
}

@inproceedings{chen2021sgpa,
  title={Sgpa: Structure-guided prior adaptation for category-level 6d object pose estimation},
  author={Chen, Kai and Dou, Qi},
  booktitle={Proceedings of the IEEE/CVF ICCV},
  pages={2773--2782},
  year={2021}
}

@inproceedings{lin2021dualposenet,
  title={Dualposenet: Category-level 6d object pose and size estimation using dual pose network with refined learning of pose consistency},
  author={Lin, Jiehong and Wei, Zewei and Li, Zhihao and Xu, Songcen and Jia, Kui and Li, Yuanqing},
  booktitle={Proceedings of the IEEE/CVF International Conference on Computer Vision},
  pages={3560--3569},
  year={2021}
}

@inproceedings{peng2022self,
  title={Self-Supervised Category-Level 6D Object Pose Estimation with Deep Implicit Shape Representation},
  author={Peng, Wanli and Yan, Jianhang and Wen, Hongtao and Sun, Yi},
  booktitle={Proceedings of the AAAI Conference on Artificial Intelligence},
  pages={2082--2090},
  year={2022}
}

@inproceedings{lunayach2024fsd,
	title={FSD: Fast Self-Supervised Single RGB-D to Categorical 3D Objects},
	author={Mayank Lunayach and Sergey Zakharov and Dian Chen and Rares Ambrus and Zsolt Kira and Muhammad Zubair Irshad},
	booktitle={Int. Conf. on Robotics and Automation},
	organization={IEEE},
	year={2024}
}

@inproceedings{fan2024acr,
  title={Acr-pose: Adversarial canonical representation reconstruction network for category level 6d object pose estimation},
  author={Fan, Zhaoxin and Song, Zhenbo and Wang, Zhicheng and Xu, Jian and Wu, Kejian and Liu, Hongyan and He, Jun},
  booktitle={Proceedings of the 2024 International Conference on Multimedia Retrieval},
  pages={55--63},
  year={2024}
}

@inproceedings{wang2024gs,
  title={Gs-pose: Category-level object pose estimation via geometric and semantic correspondence},
  author={Wang, Pengyuan and Ikeda, Takuya and Lee, Robert and Nishiwaki, Koichi},
  booktitle={European Conference on Computer Vision},
  pages={108--126},
  year={2024},
  organization={Springer}
}

@inproceedings{ikeda2024diffusionnocs,
  title={Diffusionnocs: Managing symmetry and uncertainty in sim2real multi-modal category-level pose estimation},
  author={Ikeda, Takuya and Zakharov, Sergey and Ko, Tianyi and Irshad, Muhammad Zubair and Lee, Robert and Liu, Katherine and Ambrus, Rares and Nishiwaki, Koichi},
  booktitle={2024 IEEE/RSJ International Conference on Intelligent Robots and Systems},
  pages={7406--7413},
  year={2024},
  organization={IEEE}
}

@inproceedings{tian2019fcos,
  title={Fcos: Fully convolutional one-stage object detection},
  author={Tian et al., Zhi},
  booktitle={Proceedings of the IEEE/CVF international conference on computer vision},
  pages={9627--9636},
  year={2019}
}

@article{Han2019GhostNetMF,
  title={GhostNet: More Features From Cheap Operations},
  author={Kai Han and Yunhe Wang and Qi Tian and Jianyuan Guo and Chunjing Xu and Chang Xu},
  journal={2020 IEEE/CVF Conference on Computer Vision and Pattern Recognition (CVPR)},
  year={2019},
  pages={1577-1586},
  url={https://api.semanticscholar.org/CorpusID:208310058}
}

@article{li2020generalized_QFL_DFL,
  title={Generalized focal loss: Learning qualified and distributed bounding boxes for dense object detection},
  author={Li, Xiang and Wang, Wenhai and Wu, Lijun and Chen, Shuo and Hu, Xiaolin and Li, Jun and Tang, Jinhui and Yang, Jian},
  journal={Advances in Neural Information Processing Systems},
  volume={33},
  pages={21002--21012},
  year={2020}
}

@inproceedings{lin2020convolution,
  title={Convolution in the cloud: Learning deformable kernels in 3d graph convolution networks for point cloud analysis},
  author={Lin, Zhi-Hao and Huang, Sheng-Yu and Wang, Yu-Chiang Frank},
  booktitle={Proceedings of the IEEE/CVF conference on computer vision and pattern recognition},
  pages={1800--1809},
  year={2020}
}

@inproceedings{he2017mask,
  title={Mask r-cnn},
  author={He, Kaiming and Gkioxari, Georgia and Doll{\'a}r, Piotr and Girshick, Ross},
  booktitle={Proceedings of the IEEE international conference on computer vision},
  pages={2961--2969},
  year={2017}
}

@inproceedings{yuksel2015poisson,
  title={Sample elimination for generating poisson disk sample sets},
  author={Yuksel, Cem},
  booktitle={Computer Graphics Forum},
  volume={34},
  pages={25--32},
  year={2015},
  organization={Wiley Online Library}
}

@article{xiang2017posecnn,
  title={Posecnn: A convolutional neural network for 6d object pose estimation in cluttered scenes},
  author={Xiang, Yu and Schmidt, Tanner and Narayanan, Venkatraman and Fox, Dieter},
  journal={arXiv preprint arXiv:1711.00199},
  year={2017}
}

@inproceedings{wang2021category,
  title={Category-level 6d object pose estimation via cascaded relation and recurrent reconstruction networks},
  author={Wang, Jiaze and Chen, Kai and Dou, Qi},
  booktitle={2021 IEEE/RSJ International Conference on Intelligent Robots and Systems},
  pages={4807--4814},
  year={2021},
  organization={IEEE}
}

@inproceedings{chen2021fs,
  title={Fs-net: Fast shape-based network for category-level 6d object pose estimation with decoupled rotation mechanism},
  author={Chen, Wei et al.},
  booktitle={Proceedings of the IEEE/CVF Conference on CVPR},
  pages={1581--1590},
  year={2021}
}

\end{document}